\documentclass[10pt,twocolumn,letterpaper]{article}
\usepackage[table]{xcolor}
\usepackage{cvpr}              

\usepackage[accsupp]{axessibility}  

\definecolor{cvprblue}{rgb}{0.21,0.49,0.74}
\usepackage[pagebackref,breaklinks,colorlinks,allcolors=cvprblue]{hyperref}
\usepackage{paralist}
\usepackage{indentfirst}
\usepackage{multirow}
\usepackage{makecell}
\usepackage{float}
\usepackage{xr}

\usepackage{graphicx}
\usepackage{capt-of} 
\usepackage{cuted}   

\usepackage{adjustbox}

\definecolor{rowgray1}{gray}{1.0}
\definecolor{rowgray2}{gray}{0.92}
\definecolor{rowgray3}{gray}{0.83}
\definecolor{mygreen}{RGB}{0,173,0}
\newcommand{\smallpm}{\text{\scriptsize$\pm$}}

\def\confName{CVPR}
\def\confYear{2026}

\title{Hyperbolic Multiview Pretraining for Robotic Manipulation}

\author{
Jin Yang, Ping Wei$^{\ast}$, Yixin Chen, Nanning Zheng\\
National Key Laboratory of Human-Machine Hybrid
Augmented Intelligence\\
Institute of Artificial Intelligence and Robotics, Xi'an Jiaotong University\\
{\tt\small jin.yang@stu.xjtu.edu.cn, pingwei@xjtu.edu.cn}
}

\begin{document}

\twocolumn[
{
    \maketitle
    \vspace{-0.5em}
    \centering
    \includegraphics[width=\textwidth]{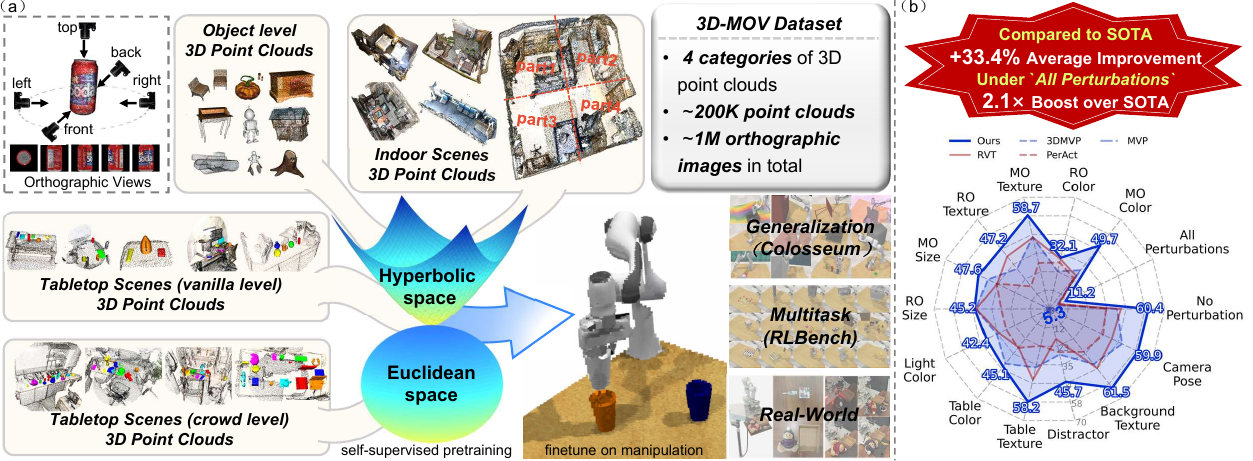}
    \captionof{figure}{Overview of HyperMVP. (a) Illustration of the HyperMVP framework, including the 3D-MOV pretraining dataset, embedding spaces, and downstream applications. (b) Comparison of generalization performance (\%) on the Colosseum~\cite{pumacay2024colosseum} under various perturbation settings. Results of HyperMVP are averaged over three evaluation runs, while other methods follow the single-run reports in~\cite{qian20253dmvp,pumacay2024colosseum}}.  
    \label{fig:introduction}
    \vspace{0.5em}
    }
]

\begin{abstract}
3D-aware visual pretraining has proven effective in improving the performance of downstream robotic manipulation tasks. However, existing methods are constrained to Euclidean embedding spaces, whose flat geometry limits their ability to model structural relations among embeddings. As a result, they struggle to learn structured embeddings that are essential for robust spatial perception in robotic applications. To this end, we propose HyperMVP, a self-supervised framework for \underline{Hyper}bolic \underline{M}ulti\underline{V}iew \underline{P}retraining. Hyperbolic space offers geometric properties well suited for capturing structural relations. Methodologically, we extend the masked autoencoder paradigm and design a GeoLink encoder to learn multiview hyperbolic representations. The pretrained encoder is then finetuned with visuomotor policies on manipulation tasks. In addition, we introduce 3D-MOV, a large-scale dataset comprising multiple types of 3D point clouds to support pretraining. We evaluate HyperMVP on COLOSSEUM, RLBench, and real-world scenarios, where it consistently outperforms strong baselines across diverse tasks and perturbation settings. Our results highlight the potential of 3D-aware pretraining in a non-Euclidean space for learning robust and generalizable robotic manipulation policies.
\end{abstract}    
\section{Introduction}
\label{sec:intro}

Developing versatile and generalizable robotic manipulation approaches is essential for real-world deployment. To this end, recent studies~\cite{shridhar2023peract,gervet2023act3d,goyal2023rvt,chen2023polarnet,goyal2024rvt2} have explored unified models for multi-task manipulation. Although they have achieved notable task-level versatility, they still exhibit weak performance under environmental perturbations, indicating limited scene generalization.

To address this limitation, recent studies~\cite{xiao2022mvp,seo2023multi,qian20253dmvp,fang2025sam2act} have drawn inspiration from the computer vision community. Some of them adopt self-supervised pretraining~\cite{xiao2022mvp,qian20253dmvp,he2022mae} on large-scale 2D or 3D non-robotic datasets to learn robust visual representations. Empirical results show that such pretraining improves the model’s capability to handle diverse manipulation tasks and scene perturbations. Despite these advances, achieving consistently strong performance across diverse tasks and scenes remains an open challenge.

Scaling up pretraining datasets improves model performance on downstream tasks, yet obtaining high-quality data remains costly. An alternative is to enhance representation quality by leveraging the geometric properties of the embedding space. We observe that existing visual pretraining methods~\cite{xiao2022mvp,qian20253dmvp,fang2025sam2act} for manipulation are typically formulated in Euclidean space with a uniform distance metric across embeddings. However, such geometric properties limit their ability to capture structured embeddings, which are essential for effective scene understanding~\cite{ge2023HCL}. To overcome this limitation, we explore visual self-supervised pretraining in non-Euclidean space and transfer its representational advantages to manipulation methods. 

We propose HyperMVP, a 3D multiview pretraining method in hyperbolic space for robotic manipulation. HyperMVP follows the pretraining–finetuning paradigm. During pretraining, we extend the MAE~\cite{he2022mae} framework with three key modifications. 1) Input design. Each 3D instance is rendered into five orthographic images to form multiview inputs. 2) Encoder design. The proposed GeoLink encoder maps Euclidean embeddings into a hyperbolic space parameterized by the Lorentz model to capture multiview hyperbolic representations. Unlike prior supervised hyperbolic representation methods~\cite{desai2023MERU,wang2025hmid,ge2023HCL}, GeoLink is trained in a fully self-supervised manner through two tailored objectives. The Top-K neighborhood rank correlation loss preserves structural and semantic consistency among patch-level image embeddings across geometric spaces. The entailment loss captures partial order relationships among embeddings. 3) Pretext tasks. We construct intra- and inter-view reconstruction tasks to facilitate self-supervised multiview pretraining. During finetuning, the pretrained GeoLink encoder is jointly optimized with the Robotic View Transformer~\cite{goyal2023rvt} (RVT) to learn manipulation policies.

Benefiting from GeoLink’s view-decoupled design, HyperMVP can scale to the arbitrary number of input views during finetuning, unlike 3D-MVP~\cite{qian20253dmvp}. Moreover, to support pretraining, we build a large-scale 3D-MOV dataset. It comprises four types of 3D point clouds, totaling about 200K instances and over 1M rendered images, as shown in Fig.~\ref{fig:introduction} (a). This facilitates analysis of how different types of 3D data affect downstream manipulation performance.

We evaluate HyperMVP on the Colosseum benchmark~\cite{pumacay2024colosseum}, which assesses manipulation generalization under diverse perturbations. As shown in Fig.~\ref{fig:introduction} (b), HyperMVP achieves a 33.4\% average improvement over the previous best baseline across all settings and attains a 2.1× performance gain in the most challenging `All Perturbations' setting. We further evaluate HyperMVP’s multi-task performance on RLBench~\cite{james2020rlbench}. Compared with training RVT from scratch or adopting models in Euclidean space, RVT with the GeoLink encoder achieves significant improvements. In addition, HyperMVP demonstrates strong real-world effectiveness while preserving comparable generalization. These results highlight the potential of hyperbolic pretraining to advance the state of the art in robotic manipulation. We also conduct ablations to analyze different design choices for pretraining.

Our key contributions are:
\begin{compactitem}
\item To the best of our knowledge, HyperMVP is the first framework to explore 3D multiview pretraining in hyperbolic space for robotic manipulation.
\item We introduce 3D-MOV, a large-scale dataset comprising four types of 3D point clouds, with each instance paired with five orthographic images. It provides a foundation for analyzing how different types of 3D data affect manipulation performance.
\item We present comprehensive evaluation results and analytical insights regarding the model’s performance across both simulated and real-world scenarios.
\end{compactitem}
\section{Related Work}
\label{sec:relatedwork}

\begin{figure*}[t]
\centering
\includegraphics[width=1\textwidth]{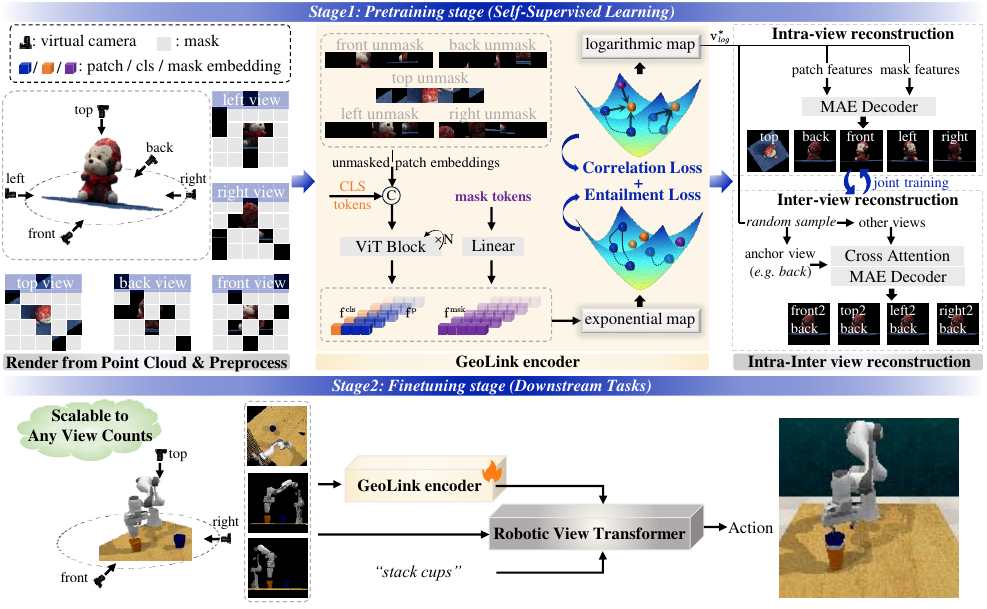}
\caption{Pipeline of HyperMVP. During pretraining, a GeoLink encoder is pretrained on multiview images rendered from point clouds. During finetuning, the pretrained GeoLink encoder is finetuned with the Robotic View Transformer for manipulation tasks.}
\label{fig:method}
\end{figure*}


\noindent \textbf{Visual Pretraining for Robotic Manipulation.} Visual pretraining has demonstrated strong generalization across diverse computer vision tasks~\cite{he2022mae,ravi2024sam2,oquab2023dinov2}, motivating its application to robotic manipulation~\cite{dasari2023unbiased,majumdar2023we,nair2023r3m,radosavovic2023real,fang2025sam2act,qian20253dmvp,zhang2024same}. These methods typically integrate a pretrained visual encoder with a visuomotor policy and jointly finetune them on manipulation tasks. HyperMVP follows the same paradigm. 

Existing visual pretraining methods for manipulation can be broadly divided into supervised and self-supervised. Supervised methods~\cite{zhang2024same,majumdar2023we,nair2023r3m,sontakke2023roboclip} typically rely on large-scale annotated 2D image or video datasets~\cite{deng2009imagenet,grauman2022ego4d,goyal2017something,miech2019howto100m}. For example, SAM2Act~\cite{fang2025sam2act,ravi2024sam2} is pretrained on over 640K masklet annotations derived from 50.9K videos and on 11M images containing more than 1B segmentation masks~\cite{kirillov2023sam}. In contrast, self-supervised methods~\cite{ma2022vip,radosavovic2023real,seo2023multi,xiao2022mvp} do not require manual annotations and are more scalable. A representative SOTA method is 3D-MVP~\cite{qian20253dmvp}, which explicitly constructs 3D multiview representations. Compared with SAM2Act, 3D-MVP is more sample-efficient, achieving competitive performance using only 200K 3D point clouds rendered into 1M multiview images. Motivated by these advantages, we explore a self-supervised multiview pretraining framework for robotic manipulation. Unlike 3D-MVP, HyperMVP learns multiview representations in hyperbolic space and is pretrained on a diverse set of point clouds.

\noindent \textbf{Hyperbolic Representation Learning.} The distance metric in Euclidean space grows linearly, which limits its ability to model hierarchical structures. In contrast, hyperbolic space expands distances exponentially, making it well suited for representing tree-like or nested structures~\cite{ganea2018hyperbolic}. Prior studies in computer vision~\cite{park2021unsupervised,desai2023MERU,ge2023HCL,wang2025hmid,mandica2024hyperbolic} and natural language processing~\cite{liu2025hyperbolic,11134045,yang2024hyperbolic} demonstrate that structured embeddings enhance both generalization and interpretability. Despite these advances, hyperbolic representation learning for robotic manipulation remains largely unexplored. Moreover, existing hyperbolic visual representation methods primarily rely on supervised signals to construct positive-negative pairs for contrastive learning~\cite{desai2023MERU,ge2023HCL,wang2025hmid}. Unlike prior methods, HyperMVP learns hyperbolic representations in a purely visual self-supervised manner. 
\section{Methodology}
\label{sec:methodology}


\noindent \textbf{Overview.} HyperMVP follows a pretraining–finetuning paradigm, as illustrated in Fig. \ref{fig:method}. During pretraining, each point cloud is rendered into five orthographic images. These images are masked~\cite{he2022mae} and fed into the GeoLink encoder to learn multiview representations in Euclidean and hyperbolic spaces. To support pretraining, we introduce intra-view and inter-view reconstruction pretext tasks and build a large-scale 3D-MOV dataset. During finetuning, the pretrained GeoLink encoder is finetuned with the Robotic View Transformer (RVT)~\cite{goyal2023rvt} to learn visuomotor policies.

\subsection{3D-MOV Dataset}
3D-MVP~\cite{qian20253dmvp} is pretrained solely on object-level point clouds, which helps the model learn object geometry. However, robotic perception extends beyond isolated objects and requires an understanding of spatial layouts and scene context. This insight prompts us to study whether pretraining on more diverse 3D data (\textit{e.g.} scene data) can improve downstream manipulation performance, and which data types are more helpful. To this end, we construct 3D-MOV, a large-scale and diverse dataset that integrates high-quality 3D data from multiple existing sources~\cite{deitke2023objaversexl,xu2022toscene,dai2017scannet}.

As illustrated in Fig.~\ref{fig:introduction} (a), we collect 180K object-level point clouds from Objaverse-XL~\cite{deitke2023objaversexl} and 1,513 indoor scenes from ScanNet~\cite{dai2017scannet}. Each indoor scene is divided into four partitions, resulting in 6,052 fine-grained scene point clouds with clearer spatial structures. Furthermore, since robotic manipulation often takes place on tabletops, we additionally include 3,999 vanilla-level and 10,001 crowd-level tabletop scene point clouds from TO-Scene~\cite{xu2022toscene}.

In total, 3D-MOV contains 200,052 high-quality 3D point clouds spanning four categories. For each point cloud, we render five orthographic RGB images (top, front, back, left, and right), resulting in about 1M multiview images. Following 3D-MVP~\cite{qian20253dmvp}, we construct a small validation set consisting of 1,000 3D point clouds for qualitative evaluation during pretraining. Because the effectiveness of pretrained representations is best assessed through their performance in downstream robotic manipulation tasks.

\subsection{Hyperbolic Preliminaries}

\noindent \textbf{Hyperbolic Space.} 
A hyperbolic space is a Riemannian manifold with constant negative curvature, unlike the flat Euclidean space of zero curvature. Among the equivalent hyperbolic models, we adopt the Lorentz model~\cite{nickel2018learning} for its stability and computational efficiency.

\noindent \textbf{Lorentz Model.} 
The Lorentz model represents an $n$-dimensional hyperbolic space by the upper sheet of a hyperboloid in $\mathbb{R}^{n+1}$. Each vector $\mathrm{x}\in\mathbb{R}^{n+1}$ is written as $\mathrm{x}=[\mathrm{x}_s,x_t]$. $\mathrm{x}_s\in\mathbb{R}^{n}$ and $x_t\in\mathbb{R}$ denote the spatial and temporal components, respectively~\cite{chen2021fully}, with $x_t=\sqrt{\left\| \mathrm{x}_s \right\| ^2+\frac{1}{c}}$. The Lorentz model $\mathcal{L} ^n$ with curvature $c$ ($c>0$) is defined as the set of all vectors that satisfy $\mathcal{L} ^n=\left\{ \mathrm{x}\in \mathbb{R} ^{n+1}:\left< \mathrm{x},\mathrm{x} \right> _{\mathcal{L}}=-1/c \right\}$, where $\left< \cdot \right> _{\mathcal{L}}$ is the Lorentzian inner product. Given $\mathrm{x},\mathrm{y}\in\mathbb{R}^{n+1}$, their Lorentzian inner product is defined as follows.
\begin{equation}
\label{eq:lorentz_inner_product}
    \left< \mathrm{x},\mathrm{y} \right> _{\mathcal{L}}=\left< \mathrm{x}_s,\mathrm{y}_s \right> -x_ty_t,
\end{equation}
where $\left< \cdot \right>$ is the Euclidean inner product.

\noindent \textbf{Geodesics.}
The geodesics in the Lorentz model are obtained as intersection curves between the hyperboloid and hyperplanes passing through the origin of $\mathbb{R}^{n+1}$. Each geodesic is the shortest path between two points on the manifold. Therefore, the Lorentz distance between two points $\mathrm{x},\mathrm{y}\in\mathcal{L}^{n}$ is given as follows.
\begin{equation}
\label{eq:geodesics}
    d_{\mathcal{L}}\left( \mathrm{x},\mathrm{y} \right) =\sqrt{1/c}\cdot \cosh ^{-1}\left( -c\left< \mathrm{x},\mathrm{y} \right> _{\mathcal{L}} \right).
\end{equation}

\noindent \textbf{Tangent Space, Exponential map, and Logarithmic map.}
The tangent space $\mathcal{T}_\mathrm{u}\mathcal{L} ^n$ at a point $\mathrm{u}\in \mathcal{L}^n$ is a Euclidean space of vectors that are orthogonal to $\mathrm{u}$ under the Lorentzian inner product (Eq.~\ref{eq:tangent_space}). In this work, $\mathrm{u}$ is fixed at the origin of the hyperboloid ($\mathrm{O}=[0,\sqrt{1/c}]$).
\begin{equation}
\label{eq:tangent_space}
    \mathcal{T}_\mathrm{u}\mathcal{L} ^n=\left\{ \mathrm{v}\in \mathbb{R} ^{n+1}:\left< \mathrm{u},\mathrm{v} \right> _{\mathcal{L}}=0 \right\}.
\end{equation}

The exponential map $\exp_{\mathrm{u}}$ is defined as a mapping from a tangent vector $\mathrm{v} \in \mathcal{T}_\mathrm{u}\mathcal{L} ^n$ to the corresponding point $\mathrm{x} \in \mathcal{L}^n$. It can be computed as follows.
\begin{equation}
\label{eq:expm}
\mathrm{x}=\exp _{\mathrm{u}}(\mathrm{v)}=\cosh \left( \sqrt{c}\left\| \mathrm{v} \right\| _{\mathcal{L}} \right) \mathrm{u}+\frac{\sinh \left( \sqrt{c}\left\| \mathrm{v} \right\| _{\mathcal{L}} \right)}{\sqrt{c}\left\| \mathrm{v} \right\| _{\mathcal{L}}}\mathrm{v},
\end{equation}
where $\left\| \mathrm{v} \right\| _{\mathcal{L}}=\sqrt{\left| \left< \mathrm{v},\mathrm{v} \right> _{\mathcal{L}} \right|}$. The logarithmic map $\log _{\mathrm{u}}$ is the inverse operation, mapping $\mathrm{x}$ from the hyperbolic space back to $\mathrm{v}$ in the tangent space. $\log _{\mathrm{u}}$ is computed as follows.
\begin{equation}
\label{eq:logm}
\mathrm{v}=\log _{\mathrm{u}}(\mathrm{x)}=\frac{\cosh ^{-1}\left( -c\left< \mathrm{u},\mathrm{x} \right> _{\mathcal{L}} \right)}{\sqrt{\left( c\left< \mathrm{u},\mathrm{x} \right> _{\mathcal{L}} \right) ^2-1}}\left( \mathrm{x}+c\mathrm{u}\left< \mathrm{u},\mathrm{x} \right> _{\mathcal{L}} \right) .
\end{equation}

\subsection{Pretraining on the 3D-MOV dataset}

\noindent \textbf{Render \& Preprocess.}
Given a 3D point cloud, we render five $224\times224$ orthographic images from multiple virtual viewpoints. Following MAE~\cite{he2022mae}, each image is divided into patches, which are mapped to patch embeddings with positional encodings. Each patch embedding is assigned a positional index in a fixed raster-scan order. A learnable view embedding is further introduced for each viewpoint and is appended to all patches of that view. This provides view identities and helps reduce inter-view ambiguity. Random masking is then applied independently to each view, and the unmasked patch embeddings are retained.

\noindent \textbf{GeoLink encoder.}
The preprocessed unmasked patch embeddings are fed into the GeoLink encoder to learn hyperbolic multiview representations. The process is described below. More details are provided in the Appendix.

\textit{\textbf{Euclidean embeddings.}} 
For each view, we prepend a learnable `CLS' token with positional encoding to aggregate global information. The resulting sequence is then processed by a stack of $N$ ViT blocks~\cite{dosovitskiy2020vit} (we set $N=8$), producing `CLS' embeddings $\mathrm{f}^\text{cls}\in \mathbb{R}^{5 \times 1 \times D}$ and patch embeddings $\mathrm{f}^\mathrm{p}\in \mathbb{R}^{5 \times P \times D}$. $P$ is the number of unmasked patches per view and $D$ is the embedding dimension. Mask embeddings $\mathrm{f}^\mathrm{msk}\in \mathbb{R}^{5 \times M \times D}$ are obtained by applying a \textit{Linear} layer to learnable mask tokens. $M$ is the number of masked patches per view. Unlike MAE~\cite{he2022mae}, which initializes mask tokens as zero vectors~\cite{he2022mae}, we randomly initialize them and apply a linear projection. It prevents mask tokens from collapsing to a single point during their mapping into hyperbolic space and ensures effective embeddings.

\textit{\textbf{Lifting Euclidean embeddings onto the hyperboloid.}} 
Once the embedding vectors $\mathrm{f}^\text{*}$ (any of $\mathrm{f}^\text{cls}$, $\mathrm{f}^\mathrm{p}$, or $\mathrm{f}^\mathrm{msk}$) are obtained in the Euclidean space, we construct the vector $\mathrm{v}^{*}=[\mathrm{f}^\text{*},0] \in \mathbb{R} ^{n+1}$. According to Eq.~\ref{eq:lorentz_inner_product} and Eq.~\ref{eq:tangent_space}, $\mathrm{v}^{*}$ belongs to the tangent space at the origin $\mathrm{O}$ of the hyperboloid, as $\left< \mathrm{v}^{*},\mathrm{O} \right> _{\mathcal{L}}=0$. We then map $\mathrm{v}^{*}$ to the hyperbolic space using the exponential map (Eq.~\ref{eq:expm}), which produces $\mathrm{x}^{*} = [\mathrm{x}_{s}^{*},x_{t}^{*}] \in \mathcal{L} ^n$. Since the temporal component of $\mathrm{v}^{*}$ is zero, only the spatial component of the Lorentz model needs to be parameterized ($\mathrm{f}^\text{*}=\mathrm{v}^{*}_{s}$). Consequently, the first term in Eq.~\ref{eq:expm} vanishes, and the computation simplifies to involve only the spatial component.
\begin{equation}
\label{eq:expm_trans_space}
\mathrm{x}_{s}^{*}=\frac{\sinh \left( \sqrt{c}\left\| \mathrm{v}^{*} \right\| _{\mathcal{L}} \right)}{\sqrt{c}\left\| \mathrm{v}^{*} \right\| _{\mathcal{L}}}\mathrm{v}^{*}_{s}=\frac{\sinh \left( \sqrt{c}\left\| \mathrm{f}^\text{*} \right\| \right)}{\sqrt{c}\left\| \mathrm{f}^\text{*} \right\|}\mathrm{f}^\text{*}.
\end{equation}
The temporal component $x_{t}^{*}=\sqrt{\left\| \mathrm{x}_{s}^{*} \right\| ^2+\frac{1}{c}}$. Moreover, to avoid numerical overflow in the exponential mapping, we follow the stabilization strategy introduced in MERU~\cite{desai2023MERU}.

\textit{\textbf{Hyperbolic representation constraints.}}
After lifting Euclidean embeddings into hyperbolic space, we design the loss $L_{\mathrm{hyper}}$ to enforce structural organization and semantic consistency among embeddings. Our goal is to learn structurally informed multiview representations that enable robust spatial perception for robotic manipulation. For this, a key requirement is to construct a meaningful semantic topology among embeddings. Prior works~\cite{desai2023MERU,ge2023HCL} have primarily pursued this via contrastive learning. They construct positive-negative pairs from supervised signals, such as image–text pairs~\cite{desai2023MERU} or scene–object pairs~\cite{ge2023HCL}. However, the GeoLink encoder operates without such supervision. Therefore, we introduce two self-supervised losses, the patch-aware Top-$K$ rank correlation loss and the entailment loss.

\textit{Patch-aware Top-$K$ rank correlation loss.}
We aim to preserve the semantic topology consistency of patch embeddings across Euclidean and hyperbolic spaces. A straightforward method is to align inter-patch distances or distance weights across spaces. However, it fails to converge due to geometric discrepancies in distance computation. To address this issue, we propose a patch-aware Top-$K$ rank correlation loss. It enforces neighborhood-order consistency between the two spaces. For each patch, we identify its Top-$K$ nearest neighbors in both spaces and minimize discrepancies in the induced rankings. This ordinal formulation emphasizes who is closer rather than how much closer, providing a geometry-agnostic constraint.

Specifically, given the patch-level Euclidean embeddings $\mathrm{f}_{i}^{\mathrm{p}}$ and hyperbolic embeddings $\mathrm{x}_{i}^{\mathrm{p}}$ from the $i$-th view, we first construct the Euclidean distance matrix $\mathrm{D}_{i}^\mathcal{E}$ and hyperbolic distance matrix $\mathrm{D}_{i}^\mathcal{L}$ among patches, as in Eq.~\ref{eq:distance}.
\begin{equation}
\label{eq:distance}
\begin{aligned}
\mathrm{D}_{i}^{\mathcal{E}}&=\left[ \left\| {\mathrm{f}_{i}^{\mathrm{p}}}_m-{\mathrm{f}_{i}^{\mathrm{p}}}_n \right\| _2 \right] _{m,n=1}^{P},
\\
\mathrm{D}_{i}^{\mathcal{L}}&=\left[ d_{\mathcal{L}}\left( {\mathrm{x}_{i}^{\mathrm{p}}}_m,{\mathrm{x}_{i}^{\mathrm{p}}}_n \right) \right] _{m,n=1}^{P},
\end{aligned}
\end{equation}
where $\left\| \cdot \right\| _2$ is the L2-norm. Each entry denotes the geometric distance between the $m$-th and $n$-th patches. We take $\mathrm{D}_{i}^\mathcal{E}$ as the anchor for neighbor definition, since the distance distribution in Euclidean space is more numerically stable. We sort $\mathrm{D}_{i}^\mathcal{E}$ in ascending order and collect the first $K$ indices for each patch to form the Top-$K$ neighbor index matrix $\pi _{i}^{K}\left( \mathrm{D}_{i}^{\mathcal{E}} \right)$, defined as follows. 
\begin{equation}
\label{eq:argsort}
\begin{aligned}
\pi _{i}^{K}\left( \mathrm{D}_{i}^{\mathcal{E}} \right) =\left[ \pi _{i}^{m}\left( 1 \right) ,\pi _{i}^{m}\left( 2 \right) ,\cdots ,\pi _{i}^{m}\left( K \right) \right] _{m=1}^{P},
\\
\left( \mathrm{D}_{i}^{\mathcal{E}} \right) _{m\pi _{i}^{m}\left( 1 \right)}\leqslant \left( \mathrm{D}_{i}^{\mathcal{E}} \right) _{m\pi _{i}^{m}\left( 2 \right)}\leqslant \cdots \leqslant \left( \mathrm{D}_{i}^{\mathcal{E}} \right) _{m\pi _{i}^{m}\left( K \right)},
\end{aligned}
\end{equation}
where $\pi _{i}^{m}\left( K \right)$ is the index of the $K$-th nearest neighbor patch of the $m$-th patch. $\left( \mathrm{D}_{i}^{\mathcal{E}} \right) _{m\pi _{i}^{m}\left( K \right)}$ denotes the Euclidean distance between the $m$-th and $\pi _{i}^{m}\left( K \right)$-th patches. 

To compare neighbor rankings rather than distances, we convert each row of $\mathrm{D}_{i}^\mathcal{E}$ and $\mathrm{D}_{i}^\mathcal{L}$ into rank positions, thereby forming neighbor-rank matrices in both spaces. This is implemented by applying \textit{argsort} twice along the last dimension of the distance matrices. We then use $\pi _{i}^{K}\left( \mathrm{D}_{i}^{\mathcal{E}} \right)$ to \textit{gather} the corresponding rank values from both neighbor-rank matrices. This results in the Top-$K$ neighbor rank matrices ${\mathrm{R}_{i}^{\mathcal{E}}}_{\pi _{i}^{K}}\in \mathbb{R}^{P\times K}$ and ${\mathrm{R}_{i}^{\mathcal{L}}}_{\pi _{i}^{K}}\in \mathbb{R}^{P\times K}$ for Euclidean and hyperbolic spaces, respectively. Finally, we define the Top-$K$ rank correlation loss $L_{\mathrm{corr}}$ as follows.
\begin{equation}
\label{eq:loss_corr}
L_{\mathrm{corr}}=1-\frac{1}{5}\sum_{i=1}^{5}{g\left( \left| {\mathrm{R}_{i}^{\mathcal{E}}}_{\pi _{i}^{K}} \right|_{\mathrm{z}} \odot \left| {\mathrm{R}_{i}^{\mathcal{L}}}_{\pi _{i}^{K}} \right|_{\mathrm{z}} \right)},
\end{equation}
where $\odot$ is the Hadamard product, $\left| \cdot \right|_{\mathrm{z}}$ is the z-score normalization, and $g(\cdot)$ is the mean operator over all elements.

\begin{figure*}[!t]
  \centering
  \includegraphics[width=1\linewidth]{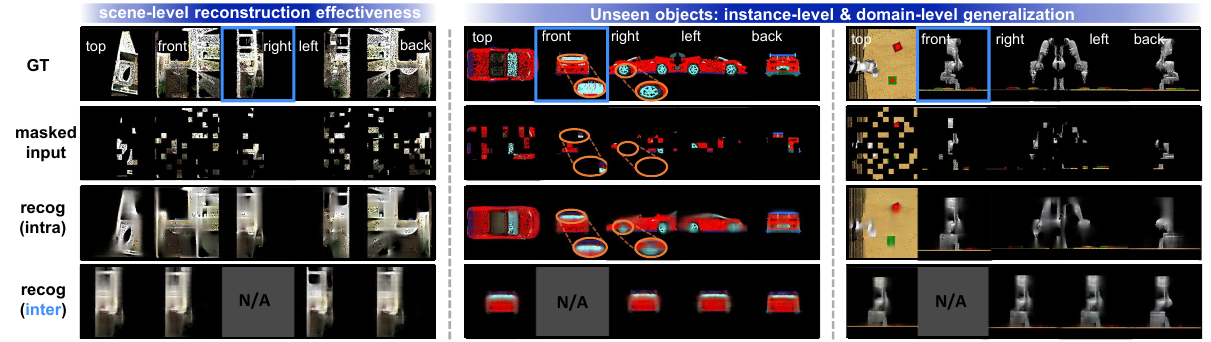}
  \caption{Qualitative reconstruction results. left: scene-level multiview inputs for validating effectiveness. middle: the unseen object within seen categories (instance-level generalization). right: the object from unseen categories (domain-level generalization).}
  \label{fig:visual_main}
\end{figure*}

\textit{Entailment loss.}
Following \cite{desai2023MERU,le2019inferring}, we adopt an entailment loss to model partial order relationships~\cite{vendrov2015order} among embeddings. An entailment cone~\cite{ganea2018cone} is defined around the hyperbolic CLS embedding $\mathrm{x}^{\mathrm{cls}}$. The CLS–patch entailment loss $L_{\mathrm{etl}}(\mathrm{x}^\text{cls}, \mathrm{x}^\mathrm{p})$ enforces patch embeddings inside the cone. It reinforces local–global semantic alignment while preserving the structural hierarchy. The CLS-mask entailment loss $L_{\mathrm{etl}}(\mathrm{x}^\text{cls}, \mathrm{x}^\mathrm{msk})$ applies the same constraint to mask embeddings. This ensures intra-view semantic consistency and improves reconstruction quality.

Overall, the hyperbolic representation constraint is defined as $L_{\mathrm{hyper}} = \lambda_\mathrm{c} L_{\mathrm{corr}}+\lambda_\mathrm{e1} L_{\mathrm{etl}}(\mathrm{x}^\text{cls}, \mathrm{x}^\mathrm{p})+\lambda_\mathrm{e2} L_{\mathrm{etl}}(\mathrm{x}^\text{cls}, \mathrm{x}^\mathrm{msk})$, where the coefficients $\lambda_\mathrm{c}$, $\lambda_\mathrm{e1}$, and $\lambda_\mathrm{e2}$ are empirically set to $1$, $0.5$, and $0.1$, respectively.

\textit{\textbf{Mapping Back to Euclidean Space.}}
Since most downstream policies operate in Euclidean space, we map the hyperbolic representation $\mathrm{x}^{*}$ back to Euclidean space for compatibility with these models. This is achieved using the logarithmic map (Eq.~\ref{eq:logm}). The corresponding feature vector $\mathrm{v}_{log}^{*}$ in Euclidean space is then computed as follows.
\begin{equation}
\label{eq:logm_trans}
\mathrm{v}_{log}^{*}=\frac{\cosh ^{-1}\left( \sqrt{c\left\| \mathrm{x}_{s}^{*} \right\| ^2+1} \right)}{\sqrt{c}\left\| \mathrm{x}_{s}^{*} \right\|}\mathrm{x}_{s}^{*}.
\end{equation}

\noindent \textbf{Intra- and Inter-View Reconstruction Tasks.}
Different from MAE~\cite{he2022mae}, we introduce intra-view and inter-view reconstruction as pretext tasks. Intra-view reconstruction uses a standard MAE decoder that reconstructs the corresponding view image from its patch and mask features. Inter-view reconstruction~\cite{weinzaepfel2022croco} aims to reconstruct the anchor view from other views. Specifically, one view is randomly selected as the anchor, and the remaining views use their patch features to predict the anchor image through cross-attention with the anchor’s patch and mask features. The attended features are then passed to an MAE decoder for final reconstruction. Both tasks are jointly optimized during pretraining.

\noindent \textbf{Pretraining Loss.}
During pretraining, we use $L_{\mathrm{hyper}}$ and a reconstruction loss $L_{\mathrm{recon}}$ for end-to-end self-supervised learning. The overall pretraining objective is defined as $L_{\mathrm{pretrain}} = L_{\mathrm{hyper}} + L_{\mathrm{recon}}$. $L_{\mathrm{recon}}$ consists of the intra-view reconstruction loss $L_{\mathrm{intra}}$ and the inter-view reconstruction loss $L_{\mathrm{inter}}$, defined as follows.
\begin{equation}
\label{eq:intra_loss_recon}
L_{\mathrm{intra}}=\frac{1}{5WH}\sum_{i=1}^{5}{\sum_{j=1}^{WH}{(\left\| I_{i}^{j}-\hat{I}_{i}^{j} \right\| _{2})^{2}}},
\end{equation}
where $W$ and $H$ are the image width and height, respectively. $I_{i}^{j}$ and $\hat{I}_{i}^{j}$ represent the ground-truth and reconstructed values of the $j$-th pixel in the $i$-th view, respectively. $\|\cdot\|_2$ is the L2-norm.
\begin{equation}
\label{eq:inter_loss_recon}
L_{\mathrm{inter}}=\frac{1}{4WH}\sum_{i=1}^{4}{\sum_{j=1}^{WH}({\left\| I_{a}^{j}-\hat{I}_{i\mapsto a}^{j} \right\| _{2})^{2}}},
\end{equation}
where $I_{a}^{j}$ and $\hat{I}_{i\mapsto a}^{j}$ denote the ground-truth and reconstructed pixel values of the anchor image, respectively. The overall reconstruction loss is formulated as $L_{\mathrm{recon}}= \lambda_\mathrm{ita}L_{\mathrm{intra}} + \lambda_\mathrm{ite}L_{\mathrm{inter}}$, where $\lambda_\mathrm{ita}$ and $\lambda_\mathrm{ite}$ are balancing hyperparameters, empirically set to $1$ and $0.5$, respectively.

\noindent \textbf{Implementation details.}
We pretrain HyperMVP for 100 epochs using AdamW optimizer, with a learning rate of $5.12\times 10^{-4}$ and a weight decay of $1\times 10^{-4}$. The batch size is set to 64, and the masking ratio is fixed at 0.75~\cite{he2022mae}. Input images are tokenized into $16\times16$ patches. The GeoLink encoder incorporates ViT blocks with a hidden dimension of 768 and 8 attention heads. For reconstruction, we adopt lightweight MAE decoders~\cite{he2022mae} with a hidden dimension of 512. Separate decoders are used for intra-view and inter-view settings, each consisting of two Transformer blocks~\cite{vaswani2017attention}. In the inter-view setting, cross-attention modules are applied with 8 attention heads and 2 layers. The model is trained on 8 NVIDIA 4090 GPUs.

\subsection{Finetuning on Manipulation Tasks}
We finetune the GeoLink encoder with RVT~\cite{goyal2023rvt} on downstream robotic manipulation tasks. During finetuning, the robot-perceived 3D scene point cloud is rendered into multiple orthographic images. These images, together with the task goal, are fed into the network for action prediction. Specifically, features extracted from GeoLink are fused with those from RVT and forwarded to the prediction layers. Notably, unlike 3D-MVP, HyperMVP flexibly scales to any number of input views during finetuning.

\noindent \textbf{Implementation details.}
We train the downstream models in both simulation (COLOSSEUM~\cite{pumacay2024colosseum}, RLBench~\cite{james2020rlbench}) and real-world environments using 8 NVIDIA RTX 4090 GPUs. The learning rate is set to $2\times10^{-3}$. The batch size is set to 20 and 16 for the simulation and real-world settings, respectively. We adopt the LAMB optimizer~\cite{you2019lamb} and train the models for 50K steps in simulation and 4K steps in real-world settings, with a 2K-step warm-up in both cases.
\section{Experiments}
\label{sec:experiments}

\subsection{Qualitative Analysis of Reconstruction}
We first validate the effectiveness and generalization of the 3D masked autoencoder in HyperMVP. Specifically, we examine whether the pretrained model can reconstruct complex scene-level inputs and generalize to unseen 3D objects. Fig.~\ref{fig:visual_main} (left) shows reconstruction results on multiview scene inputs. Despite a high masking ratio (0.75), HyperMVP successfully restores the global layout. It demonstrates the model’s strong capacity for capturing complex spatial structures. Fig.~\ref{fig:visual_main} (middle) shows reconstruction on the unseen object within seen categories (instance-level generalization). The model produces high-fidelity outputs across views and recovers fine-grained local details (\textit{e.g.}, the car's tires). Fig.~\ref{fig:visual_main} (right) presents reconstruction on objects from unseen categories (domain-level generalization). HyperMVP maintains reconstruction quality even in out-of-distribution scenarios, indicating its strong generalization capability. These results provide strong evidence that HyperMVP learns meaningful 3D representations. More visualizations and analyses are provided in the Appendix.

\begin{table*}[htbp]
    \centering
    \renewcommand{\arraystretch}{1.} 
    \caption{Multi-task performance (\%) on RLBench. Our results are reported as the mean and standard deviation over four evaluation runs.}
    \label{tab:comparison_rlbench}
    \begin{adjustbox}{max width=\textwidth}
        \begin{tabular}{lcccccccccc}
        \toprule
        \multicolumn{1}{c|}{Models} & \makecell*[c]{Pretrain [Type] / \\ Finetune [Method]} & \multicolumn{1}{c|}{\makecell*[c]{Avg. \\Success$\uparrow$}}  & \makecell*[c]{Close \\ Jar} & \makecell*[c]{Drag \\ Stick} & \makecell*[c]{Insert \\ Peg} & \makecell*[c]{Meat off \\ Grill} & \makecell*[c]{Open \\ Drawer} & \makecell*[c]{Place \\ Cups}  & \makecell*[c]{Place \\ Wine} & \makecell*[c]{Push \\ Buttons} \\
        \midrule

        \rowcolor{rowgray1}
        \multicolumn{1}{l|}{PolarNet \cite{chen2023polarnet}\scriptsize{\textit{CoRL'23}}} & - & \multicolumn{1}{c|}{46.4} & 36.0 & 92.0 & 4.0 & \textbf{100.0} & 84.0 & 0.0 & 40.0 & 96.0\\

        \rowcolor{rowgray1}
        \multicolumn{1}{l|}{PerAct \cite{shridhar2023peract}\scriptsize{\textit{CoRL'23}}} & - & \multicolumn{1}{c|}{49.4} & 55.2{\smallpm}\scriptsize{4.7} & 89.6{\smallpm}\scriptsize{4.1} & 5.6{\smallpm}\scriptsize{4.1} & 70.4{\smallpm}\scriptsize{2.0} & \textbf{88.0}{\smallpm}\scriptsize{5.7} & 2.4{\smallpm}\scriptsize{3.2} & 44.8{\smallpm}\scriptsize{7.8} & 92.8{\smallpm}\scriptsize{3.0}\\

        
        \rowcolor{rowgray1}
        \multicolumn{1}{l|}{RVT \cite{goyal2023rvt}\scriptsize{\textit{CoRL'23}}} & - & \multicolumn{1}{c|}{62.9} & 
        52.0{\smallpm}\scriptsize{2.5} & 
        99.2{\smallpm}\scriptsize{1.6} & 
        11.2{\smallpm}\scriptsize{3.0} & 
        88.0{\smallpm}\scriptsize{2.5} & 
        71.2{\smallpm}\scriptsize{6.9} & 
        4.0{\smallpm}\scriptsize{2.5} & 
        91.0{\smallpm}\scriptsize{5.2} & 
        \textbf{100.0}{\smallpm}\scriptsize{0.0} \\

        \rowcolor{rowgray2}
        \multicolumn{1}{l|}{SAM2Act \cite{fang2025sam2act}\scriptsize{\textit{ICML'25}}} & Supervised/RVT & \multicolumn{1}{c|}{68.0}& 
        71.0{\smallpm}\scriptsize{5.0} & 
        \textbf{100.0}{\smallpm}\scriptsize{0.0} & 
        13.0{\smallpm}\scriptsize{8.9} & 
        95.0{\smallpm}\scriptsize{3.8} & 
        79.0{\smallpm}\scriptsize{3.8} & 
        \textbf{9.0}{\smallpm}\scriptsize{2.0} & 
        93.0{\smallpm}\scriptsize{2.0} & 
        \textbf{100.0}{\smallpm}\scriptsize{0.0} \\

        \rowcolor{rowgray2}
        \multicolumn{1}{l|}{3D-MVP \cite{qian20253dmvp}\scriptsize{\textit{CVPR'25}}} & Self-Supervised/RVT & \multicolumn{1}{c|}{67.5} & 76.0 & \textbf{100.0} & \textbf{20.0} & 96.0 & 84.0 & 4.0 & \textbf{100.0} & 96.0\\

        \rowcolor{blue!10}
        \multicolumn{1}{l|}{\textbf{HyperMVP(Ours)}} & Self-Supervised/RVT & \multicolumn{1}{c|}{\cellcolor{blue!37}\textbf{71.1}} & 
        \textbf{80.0}{\smallpm}\scriptsize{3.3} & 
        98.0{\smallpm}\scriptsize{2.3} & 
        12.0{\smallpm}\scriptsize{5.7} & 
        97.0{\smallpm}\scriptsize{3.8} & 
        75.0{\smallpm}\scriptsize{6.0} & 
        7.0{\smallpm}\scriptsize{3.8} & 
        94.0{\smallpm}\scriptsize{5.2} & 
        \textbf{100.0}{\smallpm}\scriptsize{0.0} \\

        \midrule
        \multicolumn{1}{c|}{Models} & \makecell*[c]{Put in \\ Cupboard} & \makecell*[c]{Put in \\ Drawer} & \makecell*[c]{Put in \\ Safe} & \makecell*[c]{Screw \\ Bulb} & \makecell*[c]{Slide \\ Block} & \makecell*[c]{Sort \\ Shape} & \makecell*[c]{Stack \\ Blocks} & \makecell*[c]{Stack \\ Cups} & \makecell*[c]{Sweep to \\ Dustpan} & \makecell*[c]{Turn \\ Tap} \\
        \midrule
  
        \rowcolor{rowgray1}
        \multicolumn{1}{l|}{PolarNet \cite{chen2023polarnet}\scriptsize{\textit{CoRL'23}}}  & 12.0 & 32.0 & 84.0 & 44.0 & 56.0 & 12.0 & 4.0 & 8.0 & 52.0 & 80.0\\
        
        \rowcolor{rowgray1}
        \multicolumn{1}{l|}{PerAct \cite{shridhar2023peract}\scriptsize{\textit{CoRL'23}}} & 28.0{\smallpm}\scriptsize{4.4} & 51.2{\smallpm}\scriptsize{4.7} & 84.0{\smallpm}\scriptsize{3.6} & 17.6{\smallpm}\scriptsize{2.0} & 74.0{\smallpm}\scriptsize{13.0} & 16.8{\smallpm}\scriptsize{4.7} & 26.4{\smallpm}\scriptsize{3.2} & 2.4{\smallpm}\scriptsize{2.0} &
        52.0{\smallpm}\scriptsize{0.0} &
        88.0{\smallpm}\scriptsize{4.4} \\

        
        \rowcolor{rowgray1}
        \multicolumn{1}{l|}{RVT \cite{goyal2023rvt}\scriptsize{\textit{CoRL'23}}}& 
        49.6{\smallpm}\scriptsize{3.2} & 
        88.0{\smallpm}\scriptsize{5.7} & 
        91.2{\smallpm}\scriptsize{3.0} & 
        48.0{\smallpm}\scriptsize{5.7} & 
        81.6{\smallpm}\scriptsize{5.4} & 
        \textbf{36.0}{\smallpm}\scriptsize{2.5} & 
        28.8{\smallpm}\scriptsize{3.9} & 
        26.4{\smallpm}\scriptsize{8.2} &
        72.0{\smallpm}\scriptsize{0.0} & 
        93.6{\smallpm}\scriptsize{4.1} \\

        \rowcolor{rowgray2}
        \multicolumn{1}{l|}{SAM2Act \cite{fang2025sam2act}\scriptsize{\textit{ICML'25}}}& 
        55.0{\smallpm}\scriptsize{2.0} & 
        94.0{\smallpm}\scriptsize{4.0} & 
        89.0{\smallpm}\scriptsize{5.0} & 
        58.0{\smallpm}\scriptsize{10.6} & 
        66.0{\smallpm}\scriptsize{4.0} & 
        25.0{\smallpm}\scriptsize{3.8} & 
        \textbf{46.0}{\smallpm}\scriptsize{4.0} & 
        63.0{\smallpm}\scriptsize{2.0} &
        80.0{\smallpm}\scriptsize{8.6} & 
        88.0{\smallpm}\scriptsize{0.0} \\
        
        \rowcolor{rowgray2}
        \multicolumn{1}{l|}{3D-MVP \cite{qian20253dmvp}\scriptsize{\textit{CVPR'25}}} & \textbf{60.0} & \textbf{100.0} & 92.0 & 60.0 & 48.0 & 28.0 & 40.0 & 36.0 & 80.0 & \textbf{96.0}\\

        \rowcolor{blue!10}
        \multicolumn{1}{l|}{\textbf{HyperMVP(Ours)}}  &
        59.0{\smallpm}\scriptsize{2.0} & 
        93.0{\smallpm}\scriptsize{5.0} & 
        \textbf{97.0}{\smallpm}\scriptsize{2.0} & 
        \textbf{61.0}{\smallpm}\scriptsize{5.0} & 
        \textbf{90.0}{\smallpm}\scriptsize{5.2} & 
        26.0{\smallpm}\scriptsize{2.3} & 
        43.0{\smallpm}\scriptsize{3.8} & 
        \textbf{64.0}{\smallpm}\scriptsize{7.3} & 
        \textbf{88.0}{\smallpm}\scriptsize{0.0} & 
        \textbf{96.0}{\smallpm}\scriptsize{5.7}\\

        \bottomrule
        \end{tabular}
    \end{adjustbox}
    
\end{table*}

\subsection{Results on COLOSSEUM}
\noindent \textbf{Setup.} COLOSSEUM~\cite{pumacay2024colosseum} is a benchmark for evaluating the generalization of manipulation models. It includes 20 manipulation tasks, each with 12 single-perturbation settings and 1 challenging all-perturbation setting. The perturbations cover variations in object attributes (color, texture, size), lighting, and other factors. This benchmark closely mimics real-world scenarios and is well suited for assessing generalization under environmental changes. Following 3D-MVP~\cite{qian20253dmvp}, we generate 100 demonstrations without perturbations for each task during training. For evaluation, we construct the aforementioned perturbation settings for each task and collect 25 demonstrations per setting. 

\noindent \textbf{Metrics.} Following 3D-MVP~\cite{qian20253dmvp}, we report average success rates under each perturbation setting. 

\noindent \textbf{Results.} As shown in Fig.~\ref{fig:introduction} (b), we evaluate all baselines on COLOSSEUM under various perturbation settings. We observe that 1) HyperMVP consistently outperforms all baselines across all perturbation settings. Compared to 3D-MVP~\cite{qian20253dmvp} (previous SOTA), HyperMVP achieves a 33.4\% average improvement (35.6 $\rightarrow$ 47.5) across all perturbation settings. It indicates true robustness rather than overfitting to the training distribution. 2) HyperMVP achieves substantial improvements under all texture-related perturbations. This advantage arises from diverse data types, especially the inclusion of scene-level data. 3) Under the most challenging setting (\textit{All Perturbations}), HyperMVP achieves a 2.1$\times$ relative improvement (5.3 $\rightarrow$ 11.2). This highlights the potential of hyperbolic representations for robust manipulation. More quantitative results are provided in the Appendix.

\subsection{Results on RLBench}
\noindent \textbf{Setup.} RLBench~\cite{james2020rlbench} is a simulation-based robotic manipulation benchmark built on CoppeliaSim~\cite{rohmer2013vrep}. It defines 18 tasks with 249 variations, ranging from simple pick-and-place to high-precision operations. Each task provides 100 expert demonstrations. The environment uses a Franka Panda arm with a parallel gripper. Visual observations are captured by four RGB-D cameras (front, left shoulder, right shoulder, and wrist) at a resolution of 128×128.

\noindent \textbf{Metrics.} We present both the success rate of each task and the overall average across all tasks.

\noindent \textbf{Results.}
Tab.~\ref{tab:comparison_rlbench} reports the results on RLBench. We observe that 1) HyperMVP outperforms the supervised pretraining method (SAM2Act) and also surpasses self-supervised counterparts under the same data scale. It achieves the highest average success rate across 18 tasks. Compared with RVT trained from scratch, it delivers a 13.0\% relative improvement. These results highlight the effectiveness of self-supervised pretraining in non-Euclidean spaces for robotic manipulation. 2) HyperMVP brings limited gains on simple tasks such as \textit{Turn Tap}, while yielding substantial improvements on medium-difficulty tasks such as \textit{Stack Cups}. This is mainly because RVT already achieves a high success rate on simple tasks, leaving limited potential for further improvement. For high-precision tasks (\textit{e.g.}, \textit{Place Cups}), the improvement is marginal. Since RVT fails to solve high-precision tasks, pretraining offers little help.

\subsection{Results on Real-World}
\noindent \textbf{Setup.} We evaluate HyperMVP in real-world scenarios using a 6-DoF RealMan robot equipped with a two-finger parallel gripper and an exocentric Intel RealSense D435 depth camera. We study two manipulation tasks, a common `\textit{pick and place bear}' task and a high-precision `\textit{plug in the charging cable}' task. For each task, we collect 50 demonstrations through kinesthetic teaching and scripted execution. More details are in the Appendix.

\noindent \textbf{Metrics.} We report the success rate over ten trials.

\noindent \textbf{Results.} 
Tab.~\ref{table:real_world} presents real-world results. HyperMVP achieves an average success rate of 60.0\%, compared to 32.9\% for RVT across all task variants. We further evaluate model generalization under perturbations such as `Light Change', `MO Texture'~\cite{pumacay2024colosseum}, `Distractor'~\cite{pumacay2024colosseum}, and their combination. When all perturbations are applied, RVT suffers a 77.8\% relative drop, whereas HyperMVP shows a smaller 44.4\% decrease. In the high-precision task (b), 
RVT completely fails (0\% success rate), while stage-wise analysis shows only 20\% success in `Cable Grasping' compared to 90\% for HyperMVP. It indicates that HyperMVP improves high-precision manipulation, though the final success remains limited by the downstream visuomotor policy.

\begin{table}[tbp]\small
    \centering
    \renewcommand{\arraystretch}{1.}
    \setlength\tabcolsep{1.5pt}
    \caption{Real-world results.}
    \label{table:real_world}
    \scalebox{1.0}{
    \begin{tabular}{cc|cc}
    \toprule
    \multicolumn{1}{c}{\multirow{1}{*}{Task}} & 
    \multicolumn{1}{c|}{\multirow{1}{*}{Variants}}&
    \multicolumn{1}{c}{\makecell*[c]{RVT}}&
    \multicolumn{1}{c}{\makecell*[c]{HyperMVP}}\\
    \midrule

    \multirow{5}{*}{\makecell*[c]{(a) pick and \\ place bear}} & 
    \multicolumn{1}{l|}{No Perturbation} &
    \multicolumn{1}{c}{9/10}  & 
    \multicolumn{1}{c}{9/10} \\
    \cline{2-4}
    
    {} &\multicolumn{1}{l|}{Light Change}  &\multicolumn{1}{c}{3/10 ($\downarrow66.7\%$)} &\multicolumn{1}{c}{\textbf{5/10 ($\downarrow44.4\%$)}}\\
    
    {} &\multicolumn{1}{l|}{MO Texture}  &\multicolumn{1}{c}{4/10 ($\downarrow55.6\%$)} &\multicolumn{1}{c}{\textbf{6/10} ($\downarrow33.3\%$)}\\

    {} &\multicolumn{1}{l|}{Distractor}  &\multicolumn{1}{c}{3/10 ($\downarrow66.7\%$)} &\multicolumn{1}{c}{\textbf{6/10} ($\downarrow33.3\%$)}\\

    {} &\multicolumn{1}{l|}{All Perturbations}  &\multicolumn{1}{c}{2/10 ($\downarrow77.8\%$)} &\multicolumn{1}{c}{\textbf{5/10} ($\downarrow44.4\%$)}\\
    \midrule

    \multirow{2}{*}{\makecell*[c]{(b) plug in the \\ charging cable}} & 
    \multicolumn{1}{l|}{Cable Grasping} &
    \multicolumn{1}{c}{2/10}  & 
    \multicolumn{1}{c}{\textbf{9/10}} \\
    
    {} &\multicolumn{1}{l|}{Plug Insertion}  &\multicolumn{1}{c}{0/10} &\multicolumn{1}{c}{\textbf{2/10}}\\
    \bottomrule
    \end{tabular}
    }
\end{table}

\subsection{Ablation Studies}
To analyze the effect of various design choices in HyperMVP, we conduct ablation studies on the RLBench benchmark. Tab.~\ref{table:ablations} summarizes the results under different variants. More ablations are presented in the Appendix.

\noindent \textbf{Why not Multiview Transformer (MVT) in HyperMVP?} 
MVT~\cite{qian20253dmvp,goyal2023rvt} is widely used for multiview learning. However, it is not always optimal for large-scale pretraining, where efficiency and memory usage become critical. As shown in Group I.1, adopting MVT in HyperMVP leads to \textit{OOM} under our settings. The issue arises from the quadratic complexity of its attention operation with respect to the input sequence length. Reducing the batch size alleviates this issue but greatly slows training. Moreover, MVT enforces the same number of input views during pretraining and finetuning, which limits scalability. Therefore, we extend MAE~\cite{he2022mae} to enable efficient and flexible pretraining.

\noindent \textbf{Does pretraining in hyperbolic space truly help?} To answer this, we construct a variant MAE$^*$, which directly uses Euclidean features from the ViT blocks of the GeoLink instead of hyperbolic ones. All other settings remain identical to HyperMVP. As shown in Group I.2 (68.22 \textit{vs.} 71.11), hyperbolic representations indeed boost manipulation performance. We expect this finding to inspire future exploration of non-Euclidean representation pretraining for robotic applications. More results are provided in the Appendix.

\begin{table}[tbp]\small
    \centering
    \renewcommand{\arraystretch}{1.}
    \setlength\tabcolsep{1.5pt}
    \caption{Ablations on the RLBench benchmark. For each variant, we report the average success rate (\%) over four evaluation runs. \textit{OOM}: out of memory. \textit{w/o.}: without. }
    \label{table:ablations}
    \scalebox{1.0}{
    \begin{tabular}{cc|c|c}
    \toprule
    \multicolumn{1}{c}{\multirow{1}{*}{Group}} & 
    \multicolumn{1}{c|}{\multirow{1}{*}{Index}} & 
    \multicolumn{1}{c|}{\multirow{1}{*}{Variants}}&
    \multicolumn{1}{c}{\makecell*[c]{Average \\Success Rate}}\\
    \midrule

    \rowcolor{rowgray2}
    \multirow{1}{*}{} &{} &\multicolumn{1}{l|}{HyperMVP}  &\multicolumn{1}{c}{71.11} \\
    \midrule
    
    \multirow{3}{*}{\makecell*[c]{I \\ Pretraining \\ Network}} &1 &\multicolumn{1}{l|}{MVT from 3D-MVP~\cite{qian20253dmvp}}  &\multicolumn{1}{c}{\textit{OOM}} \\
    
    {} &2 &\multicolumn{1}{l|}{MAE$^*$}  &\multicolumn{1}{c}{68.22} \\

    {} &3 &\multicolumn{1}{l|}{MAE$^*$ \textit{w/o.} inter-view}  &\multicolumn{1}{c}{68.17} \\
    
    \midrule
    \multirow{2}{*}{\makecell*[c]{II \\ Dataset Scale}} &1 &\multicolumn{1}{l|}{\textit{w/o.} ScanNet~\cite{dai2017scannet}, $\sim$194K}  &\multicolumn{1}{c}{65.06} \\
    
    
    {} &2 &\multicolumn{1}{l|}{\textit{w/o.} TO-Scene~\cite{xu2022toscene}, $\sim$186K}  &\multicolumn{1}{c}{68.44} \\
    \midrule
    \multirow{4}{*}{\makecell*[c]{III \\ Loss}} &1 &\multicolumn{1}{l|}{\textit{w/o.} $L_{\mathrm{inter}}$}  &\multicolumn{1}{c}{71.00} \\

    {} &2 &\multicolumn{1}{l|}{\textit{w/o.} $L_{\mathrm{corr}}$}  &\multicolumn{1}{c}{67.72} \\
        
    {} &3 &\multicolumn{1}{l|}{\textit{w/o.} $L_{\mathrm{etl}}(\mathrm{x}^\text{cls}, \mathrm{x}^\mathrm{p})$}  &\multicolumn{1}{c}{70.06} \\
    
    {} &4 &\multicolumn{1}{l|}{\textit{w/o.} $L_{\mathrm{etl}}(\mathrm{x}^\text{cls}, \mathrm{x}^\mathrm{msk})$}  &\multicolumn{1}{c}{69.78} \\
    \bottomrule
    \end{tabular}
    }
\end{table}

\noindent \textbf{Which types of scene point clouds are more beneficial?} Recall that 3D-MOV is introduced to integrate diverse scene data for enhanced representation learning. To examine how different scene data contribute to this goal, we conduct an ablation study. As shown in Group II, removing ScanNet~\cite{dai2017scannet} causes a larger performance drop than removing TO-Scene~\cite{xu2022toscene}. It indicates that realistic scene data with rich structural layouts provide greater benefits. Moreover, as observed in Group II, the 194K subset performs worse than the smaller 186K subset. This result indicates that data diversity outweighs data scale. 

\noindent \textbf{Loss Ablations.} We conduct ablations on different loss components to evaluate their contributions. 1) Comparing III.1 with HyperMVP shows that the inter-view reconstruction task brings marginal gains. We argue that orthographic projection ensures geometric consistency across views, reducing the additional benefit of inter-view reconstruction. 2) Removing $L_{\mathrm{corr}}$ leads to a notable performance drop, highlighting the importance of structural regularization in hyperbolic space. 3) III.3 and III.4 show that disabling the entailment loss causes a slight performance drop, similar to the findings reported in MERU~\cite{desai2023MERU}. However, its inclusion remains essential for better interpretability.
\section{Conclusion}
\label{sec:conclusion}
We propose HyperMVP, the first hyperbolic multiview self-supervised pretraining framework for robotic manipulation. It aims to improve both the performance and generalization of manipulation methods. A principled GeoLink encoder is introduced to capture hyperbolic multiview representations and is then finetuned with visuomotor policies. We evaluate HyperMVP’s effectiveness and generalization across multiple simulation benchmarks (COLOSSEUM and RLBench) and real-world scenarios. Under the most challenging \textit{All Perturbations} setting, HyperMVP substantially outperforms the baselines. Ablation studies show that factors such as dataset diversity, embedding geometry, and loss design all affect downstream performance. Overall, these results demonstrate that integrating 3D-aware pretraining with non-Euclidean geometry helps build robust and generalizable robotic manipulation systems.

{
    \small
    \bibliographystyle{ieeenat_fullname}
    \bibliography{main}
}


\end{document}